\documentclass{article}

\usepackage{PRIMEarxiv}
\usepackage{amssymb}
\usepackage{times}
\usepackage{epsfig}
\usepackage[utf8]{inputenc} 
\usepackage[T1]{fontenc}    
\usepackage{hyperref}       
\usepackage{url}            
\usepackage{booktabs}       
\usepackage{amsfonts}       
\usepackage{nicefrac}       
\usepackage{microtype}      
\usepackage{lipsum}
\usepackage{graphicx}
\usepackage{amsmath}
\usepackage{amssymb}
\usepackage{physics}
\usepackage{multirow} 
\usepackage{caption}
\usepackage{subcaption}
\usepackage[dvipsnames]{xcolor}
\graphicspath{{media/}}     

\title{Focus-and-Detect: A Small Object Detection Framework for Aerial Images
\thanks{\textit{\underline{Citation}}: 
\textbf{Koyun, Onur Can, Reyhan Kevser Keser, İbrahim Batuhan Akkaya, and Behçet Uğur Töreyin. "Focus-and-Detect: A small object detection framework for aerial images." Signal Processing: Image Communication (2022): 116675, DOI:10.1016/j.image.2022.116675}} 
}

\author{
  Onur Can Koyun \\
  İstanbul Technical University \\
  İstanbul, Turkey\\
  \texttt{okoyun@itu.edu.tr} \\
   \And
  Reyhan Kevser Keser \\
  Istanbul Technical University \\
  İstanbul, Turkey\\
  \texttt{keserr@itu.edu.tr} \\
  \And
  İbrahim Batuhan Akkaya\\
  Aselsan\\
  Ankara, Turkey\\
  \texttt{ibakkaya@aselsan.com.tr} \\
  \And
  Behçet Uğur Töreyin \\
  Istanbul Technical University \\
  İstanbul, Turkey\\
  \texttt{toreyin@itu.edu.tr} \\
}

\begin{document}
\maketitle

\begin{abstract}
Despite recent advances, object detection in aerial images is still a challenging task. Specific problems in aerial images makes the detection problem harder, such as small objects, densely packed objects, objects in different sizes and with different orientations. To address small object detection problem, we propose a two-stage object detection framework called "Focus-and-Detect". The first stage which consists of an object detector network supervised by a Gaussian Mixture Model, generates clusters of objects constituting the focused regions. The second stage, which is also an object detector network, predicts objects within the focal regions. Incomplete Box Suppression (IBS) method is also proposed to overcome the truncation effect of region search approach. Results indicate that the proposed two-stage framework achieves an AP score of 42.06 on VisDrone validation dataset, surpassing all other state-of-the-art small object detection methods reported in the literature, to the best of authors' knowledge.
\end{abstract}


\section{Introduction}

\begin{figure*}[ht!]
\begin{center}
   \includegraphics[width=1\linewidth]{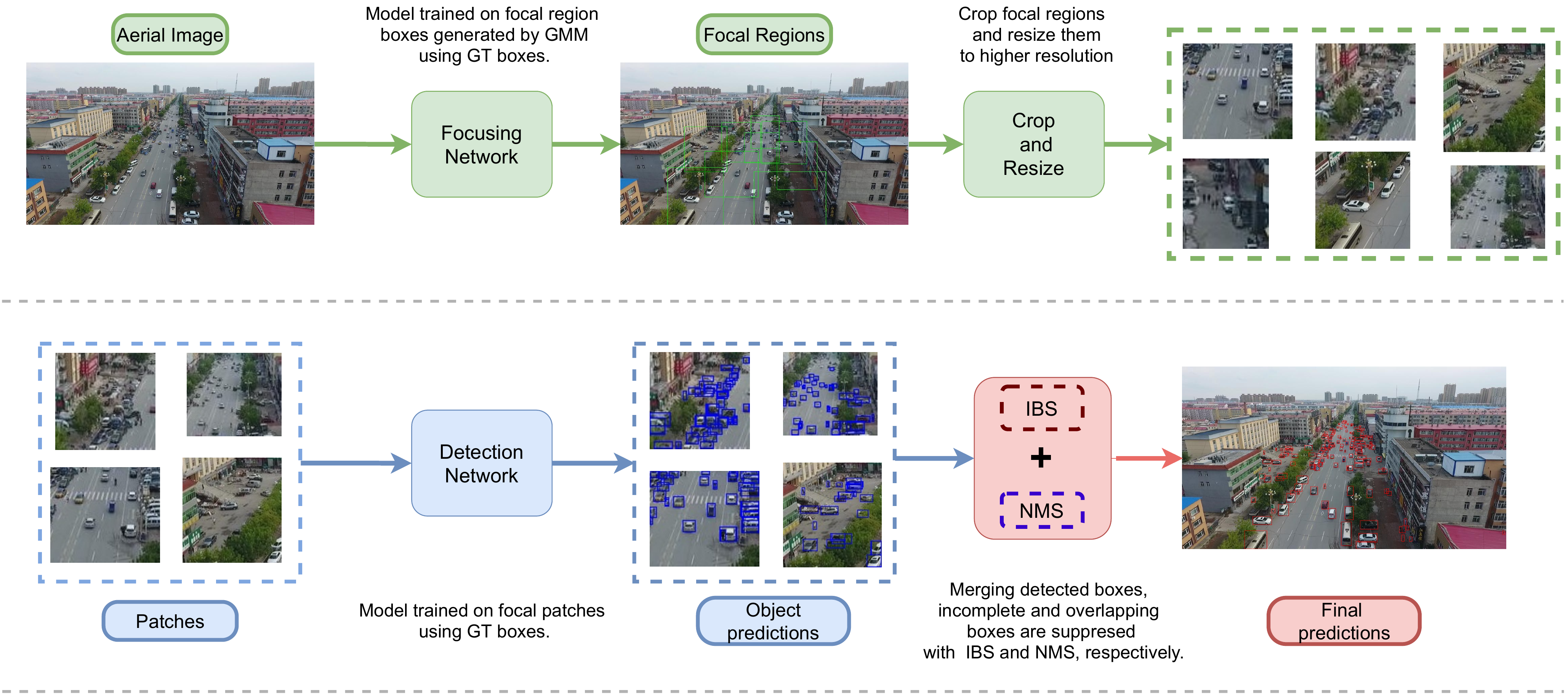}
\end{center}
   \caption{Focus\&Detect (F\&D) framework. F\&D consists of two components : (1) Focus network; (2) Detection network. While focus network detects the possible object containing regions (Focal Regions), detection network detects objects in these focal regions. Final predictions are generated by merging the predictions of focal regions. NMS and IBS methods are applied to eliminate overlapping and truncated boxes. Both detectors are trained with supervision. Focus network utilizes cluster coordinates generated by a Gaussian Mixture Model as supervision signal. On the other hand, detection network utilizes object ground truth bounding boxes in each respective focal region.
    }
\label{fig:m2}
\end{figure*}
Object detection is a computer vision task which consists of two sub-tasks, namely, object localization and classification. It is one of the fundamental problems, since many other tasks rely on it, such as image captioning, object tracking, instance segmentation and scene understanding \cite{tong2020recent}. Thus, it has been studied for a long time. With the progress of deep learning based methods, handcrafted feature based methods, such as HOG \cite{dalal2005histograms} and SIFT \cite{lowe1999object}, have become obsolete. \textcolor{black}{ SIFT and HOG features are low-level features which cannot be utilized as hierarchical layer-wise representations while the deep models are able to represent the data as hierarchical combination of abstract representations.} Nevertheless, recent methods are getting more complex day by day thanks to the development on hardware capabilities. In \cite{liu2021survey}, deep learning based methods are defined as a combination of various components. \textcolor{black}{In general, detection networks consist of backbone, neck and head. In this context, backbone model is the network that extracts features for the detection task, head is the actual detection model that predicts both bounding boxes and classes, neck is placed between backbone and head networks and fuses feature maps from different stages of backbone model. There are different approaches for detection heads, such as one-stage detection and two-stage detection models. One-stage detection models do not include a region proposal layer \cite{ren2015faster} in the head model and run detection directly over a dense sampling of locations. On the other hand two-stage models utilize region proposal network to extract object regions which are used for bounding box regression and classification.}


Aerial object detection, which can be categorized as a case of the general small object detection problem, is an emerging field with recent advances. Although, it has a wide range of applications, such as surveillance, precision agriculture, military monitoring and urban management \cite{butte2021potato, walha2013moving}, it is one of the most challenging computer vision tasks. 
Earlier, several studies proposed adapting methods established for natural images to aerial images \cite{zhang2016weakly,cheng2016rifd}. However, various difficulties arose due to such an approach \cite{xia2018dota}. First, in aerial images, orientation and aspect ratios may substantially differ from natural images. Second, scale variation is much severe in aerial images for both intra-class and inter-class samples \cite{wang2019spatial}. As an example, \cite{zhou2019scale} reports the statistics for `car' class in the MS COCO and the VisDrone \cite{zhu2018vision} datasets. It is indicated that in the VisDrone dataset, the variance of the sizes of `car' objects is almost five times larger than that of the MS COCO dataset. Third, objects in aerial images are small and densely placed. For example, up to 902 objects may exist in a single image in VisDrone Detection dataset \cite{zhu2018visdrone}. Moreover, class imbalance problem exists in aerial images \cite{zhu2018visdrone}, where it makes the small object detection problem even harder for classes with small number of samples. Hence, dedicated approaches addressing the aforementioned problems are required for the small object detection task.

Region search is a powerful method for small object detection, which aims to find and focus on regions that potentially include objects \cite{yang2019clustered, wang2020object}. Since aerial images consist of dense and small objects, we focus on region search for aerial object detection problem, in this paper. For this purpose, we propose a framework consisting of two stages, namely the focus and the detection stages.
In the first stage, regions to be focused are determined by a detector which is supervised by a Gaussian Mixture Model. The second stage, fed by these regions which are mainly clusters of objects, predicts objects within these regions. While merging the predictions on these regions, NMS and the proposed IBS methods are utilized to eliminate overlapping and truncated bounding boxes.

Our contributions can be listed as follows:
\begin{itemize}
    \item We propose a framework, namely, `Focus\&Detect' for small object detection in aerial images which is based on region search
    \textcolor{black}{\item We propose a method to generate object clusters using Gaussian mixture model, where the generated clusters are scale normalized.}
    \item We also propose the `Incomplete Box Suppression' (IBS) approach to suppress incomplete boxes caused by overlapping focal regions.
    \item Our proposed method achieves 42.06 AP score on the VisDrone validation set and \textcolor{black}{54.16 AP@70 score on UAVDT test set}. To the best of our knowledge, our method outperforms state-of-the-art methods for small object detection, reported in the literature on VisDrone dataset.
\end{itemize}

\section{Related Work}

In this section we briefly review related work in the directions of object detection, small object detection and object detection in aerial images.

\subsection{\textcolor{black}{ Object Detection}}
\textcolor{black}{Recent methods in object detection literature typically employ powerful models for backbone, such as ResNet~\cite{he2016deep}, Hourglass~\cite{newell2016stacked} and ResNeXt~\cite{xie2017aggregated}. Feature pyramid network \cite{lin2017feature} based architectures are the main choice for neck model. There are multi-stage head models in the literature such as Faster-RCNN \cite{ren2015faster}, Mask-RCNN  \cite{he2017mask}, Cascade-RCNN  \cite{cai2018cascade}. Faster R-CNN generates proposals by the region proposal network (RPN). Mask R-CNN extends Faster R-CNN to perform detection and segmentation tasks simultaneously. On the other hand, YOLOv3  \cite{redmon2018yolov3}, SSD  \cite{liu2016ssd}, GFL  \cite{li2020generalized}  and RetinaNet  \cite{lin2017focal} are examples of single-stage detectors. Single-stage detectors omit the proposal stage and make detection on the dense sample of locations.}

Recent two-stage methods include RPN \cite{ren2015faster} based modules to generate region proposals. Anchor is an alternative concept in object detection literature standing for pre-defined bounding boxes that are to be matched with ground-truth bounding boxes of objects. Some studies propose anchor-free approaches to avoid the computational cost of using anchors \cite{tian2019fcos, duan2019centernet}. A key component for deep learning based object detection methods is the loss function. It mainly consists of two terms corresponding to the regression and the classification losses \cite{ren2015faster}. Regression and classification branches are the final modules of an object detector, which predict the localization and classification of objects, respectively. Finally, Non-Maximum Suppression (NMS) and its variations \cite{bodla2017soft, neubeck2006efficient} are instrumental in the workflow of an object detector, since object detectors typically generate lots of redundant predictions and NMS is used for reducing redundancy. With the improvements on these components, great progress has been made on generic object detection \cite{dai2021dynamic, liang2021cbnetv2}, whereas small object detection still needs improvements to obtain satisfying detection performances \cite{liu2021survey}. Adapting an object detector which performs adequately well for medium or large objects, cannot yield sufficient performances on small objects, whose areas are less than $32 \times 32$ pixels as stated in \cite{lin2014microsoft}. For instance, it is indicated that recently proposed DETR's \cite{carion2020end} detection performance is not at the desired level for small objects, whereas DETR performs better than Faster R-CNN \cite{ren2015faster} on MS COCO \cite{lin2014microsoft} dataset.

\subsection{Small Object Detection}
Small object detection task is an important computer vision problem with applications in various fields, such as autonomous driving, UAV-based imaging, and surveillance. Even though, it is a crucial tool for different computer vision tasks in numerous fields, performances of the current methods are not at the desired level. Still, most of the object detection methods struggle with small objects due to the issues, such as the inadequacy of information raised from the area covered by small objects on the image, high possibility of location for small objects and being adapted for medium and large objects \cite{tong2020recent}.

To solve the inadequacy of information problem for small object detection, \cite{noh2019better} uses super-resolution techniques to improve the performance of Faster R-CNN, whereas \cite{rabbi2020small} utilizes a super-resolution GAN on remote sensing images. 

Increasing the resolution of input yield better performance for small objects. Thus, some trivial methods are also proposed such as using an image pyramid as the input to improve the performance of detecting small faces \cite{zhang2019fast}. However these methods are not scalable efficiently.

In \cite{marvasti2020comet}, a two-stream network is proposed which utilizes multi-scale representation as well as the attention mechanism. Another study \cite{sun2021mask} uses contextual information besides the multi-scale representation obtained from SSD model. Pan et. al.\cite{pan2020tdfssd} proposed a  multi-scale feature fusing scheme to improve small object detection on SSD model. In another study \cite{yin2021fd}, SSD model is revised with feature fusion and dilated convolutions increasing the detection performance. 

On the other hand, some methods are focused on improving the region-proposal stage using various techniques such as advancing anchors \cite{eggert2017improving} and increasing samples of small objects \cite{kisantal2019augmentation}.  In \cite{dai2019hybridnet}, authors proposed a hybrid model which uses both region-proposal and dense detection heads to increase performance.

\subsection{Object Detection in Aerial Images}

Aerial images constitute one of the most difficult cases for object detection as they mostly comprise small objects, large difference between number of samples of different classes and the high scale variance on both of inter-class and intra-class. To alleviate these difficulties, numerous methods are proposed previously. For example,  an adaptive augmentation method is proposed for the class imbalance problem in \cite{chen2019rrnet}, which is called AdaResampling. In \cite{hong2019patch}, a hard chip mining method is proposed as data augmentation on aerial images. Moreover, \cite{wang2019spatial} proposes an improvement on obtaining multi-scale features in order to reduce the effect of scale variance for object detection.

Since aerial images mostly consist of small and dense objects, some methods focus on improving region search \cite{hong2019patch, li2020density,  ozge2019power, tang2020penet, wang2020object, wei2020amrnet, yang2019clustered, zhang2019fully}. For example, \cite{ozge2019power} proposes tiling based method to detect pedestrians and vehicles in aerial images in real-time.   
In \cite{wang2020object}, difficult cluster regions are determined using mean shift algorithm to feed the object detector.
\cite{wei2020amrnet} proposes three augmentation methods for cropping based approach which are mosaic augmentation, adaptive cropping and mask resampling.
In \cite{zhou2019scale}, an adaptive image cropping method based on FPN \cite{lin2017feature} is proposed to solve scale challenges in aerial images.\cite{li2020density} constructs density maps to determine regions to be cropped. Then an object detector is fed by these crops as well as the whole image.
\cite{yang2019clustered} utilizes clustering to obtain image crops. Before feeding object detector with these crops, this method re-scales them by determining the proper scales for the objects, to avoid the degradation on performance. \textcolor{black}{Contrary to  \cite{li2020density,yang2019clustered}, our method utilizes predicted regions only, and does not employ detection on the whole image. Furthermore, \cite{li2020density} uses density maps to generate regions of interest, which does not directly normalize the object scales. \cite{yang2019clustered} utilizes a sub-network to predict scales of detected clusters which means additional computation. On the other hand, Gaussian mixture model provides scale normalization across predicted regions without additional computation as resizing predicted regions to a fixed size, yields a shift in the mean of each mixture component and resulting normalization of bounding boxes.}

\begin{figure}[t!]
\begin{center}
   \includegraphics[width=1\linewidth]{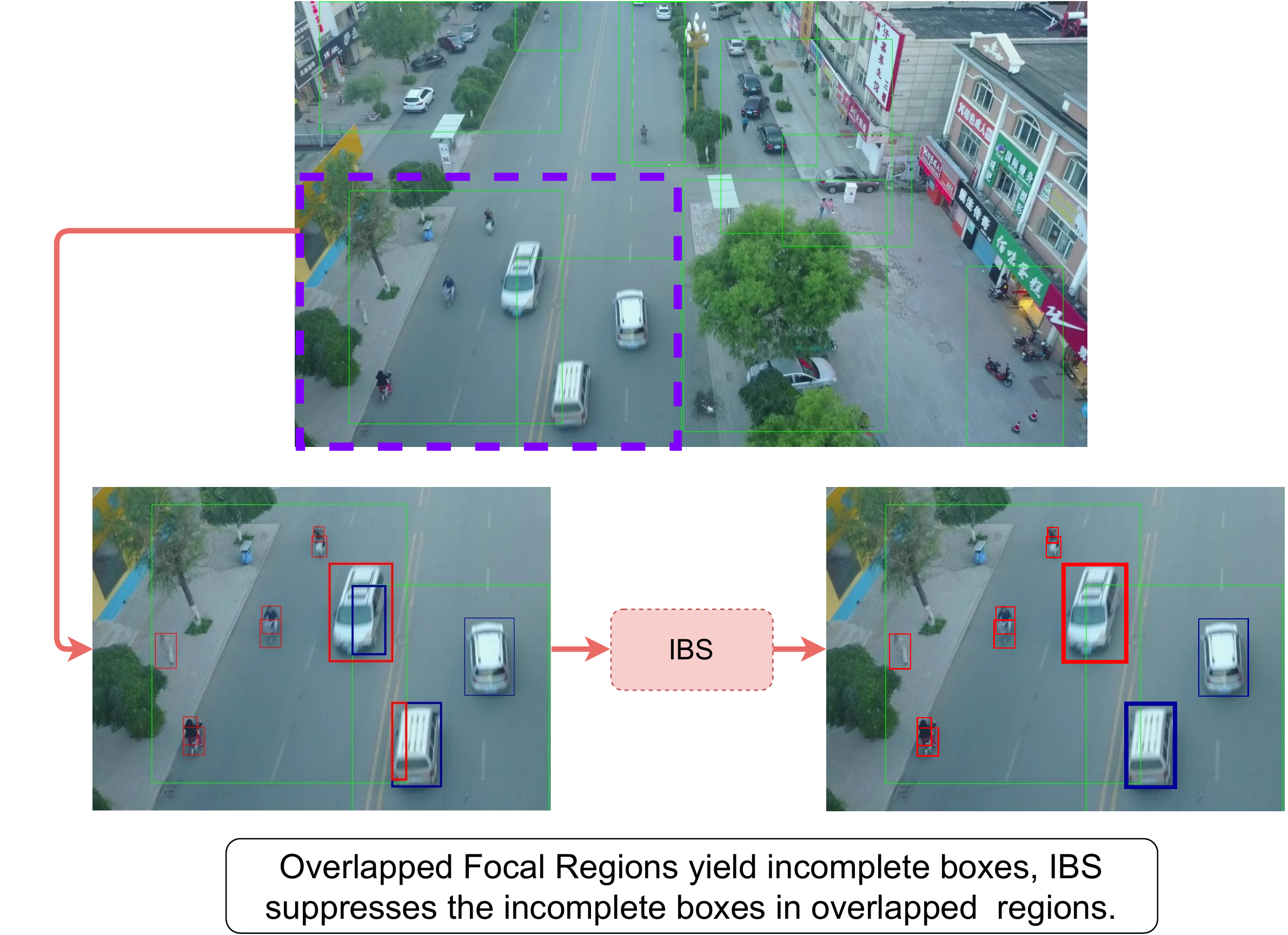}
\end{center}
   \caption{Focus-and-Detect predicts focal regions comprising object clusters, and detects objects in these regions. Merging predictions of all focal regions constitute the final step. Overlapping focal regions might yield incomplete bounding box predictions that fail to fully cover the whole areas of objects. The blue box contained within the red one and the red box within the blue one in the lower left subfigure are examples of such incomplete boxes. Incomplete boxes cause wrong predictions for object class affiliations. To overcome this issue, we proposed a method called Incomplete Box Suppression (IBS), where each box prediction in a focal region is able to suppress predictions of other focal regions, yielding only the `complete' predictions.}
\label{fig:m1}
\end{figure}

Different from the previous studies, we propose to use Gaussian Mixture Model (GMM) for region search.  Moreover, we propose Incomplete Box Suppression (IBS) in order to suppress incomplete boxes within overlapping regions that are generated by the first detector under supervision of GMM. Figure~\ref{fig:m1} demonstrates the contribution of proposed IBS method. 

\section{Focus-and-Detect}
\subsection{Overview}
In general, object detection performance on aerial images are hindered by the small objects, changes in the perspective of objects, occlusion and truncation. Using high resolution images as input is one of the simplest solutions to the small object detection problem. Unfortunately, high-resolution images impose an unaffordable computational cost to deep neural networks.  Using a focusing mechanism and increasing the resolution of the focal regions have the advantages of this simple method, but at lower computational cost. As shown in Figure~\ref{fig:m2}, detection on aerial images consist of two stages: Focus network which detects focal regions consisting of cluster of objects, detection network which detects objects in focal regions. Post processing methods are applied after merging the predictions. Specifically, we proposed the Incomplete Box Suppression (IBS) mechanism to suppress incomplete boxes from overlapping focal regions. We also use standard non-max suppression (NMS) to suppress overlapping boxes after merging the predicted boxes.

\subsection{Focus Stage}
Focus stage consists of an object detection network, trained to detect focal regions. Focal regions are generated using a Gaussian Mixture Model via ground truth bounding boxes. Generalized focal loss (GFL) \cite{li2020generalized} is selected as base detection method. Backbone of the model is ResNet-50 network with deformable convolutional layers \cite{zhu2019deformable}. 

Second part of model, namely, Feature Pyramid Network (FPN) aims to exploit and refine the feature maps obtained from different stages of ResNet-50, and the last part is detection head of the model which predicts bounding boxes of focal regions. The deformable convolution  layer is used in the last
three-stage  of the backbone. 

The traditional convolutional network has limited performance on geometric transformation due to the restricted form of convolutional layers and pooling layers \cite{dai2017deformable}. The traditional network
architecture is not able to transfer well on the focal region
detection task. Transferability of focal region features  are inferior to transferability of traditional object features. In order to improve the transferability of the learned features, the deformable convolutional layers \cite{zhu2019deformable} has been utilized within ResNet-50, since deformable convolution can change
the reception field dynamically. The proposed change leads to better representation of focal regions.

The performance of the overall framework mostly depends on the focus stage. Ideally, predicted focal regions must include all the object bounding boxes without any truncation. However, there might be overlapped regions and truncated objects in these regions. These issues are resolved by employing the IBS method as a post-processing stage, and presented in detail below.

\subsubsection{Generating Ground-Truth Boxes of Focal Regions Using Gaussian Mixture Model }

\textcolor{black}{In object localization problem, areas of objects in the same class can be modeled with a Gaussian distribution, as object sizes do not vary much. This assumption is true for object detection datasets such as MS COCO \cite{lin2014microsoft} or PASCAL VOC \cite{everingham2010pascal}. However, in aerial image datasets such as VisDrone \cite{zhu2018visdrone}, object areas deviate from one to another depending on the angle and the altitude of the camera. Instead of a single Gaussian model, a Gaussian mixture model is a better choice whereas, contrary to a single Gaussian model, a mixture model consist of Gaussians with smaller deviations when object locations are used as input to the mixture model. }

\begin{figure}[h!]
\begin{center}
   \includegraphics[width=0.6\linewidth]{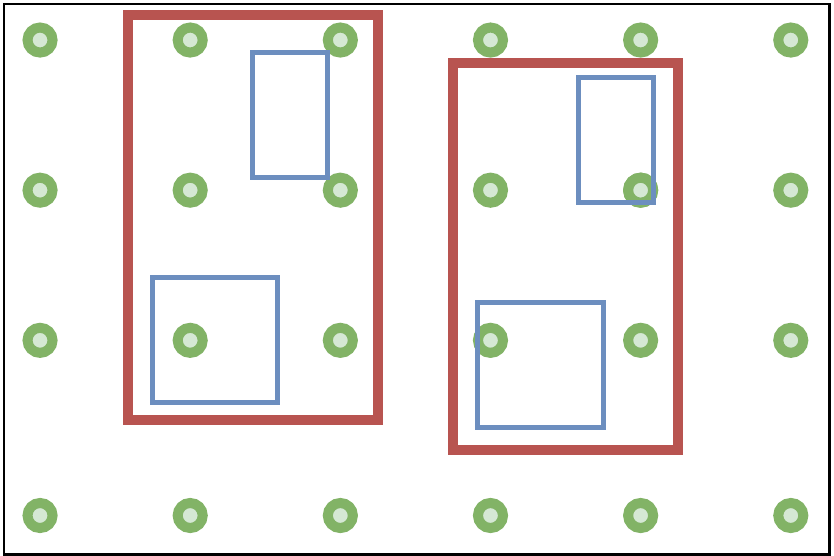}
\end{center}
   \caption{ \textcolor{black}{Distance vectors are defined as vectors emanating from the predefined grid of points and ending at the center of the ground truth bounding box coordinates. Distance vectors are then input to GMMs for clustering bounding boxes at each image. }\textcolor{blue}{Blue boxes} represent GT boxes of objects, \textcolor{black}{red boxes} represent generated focal regions.
    }
\label{fig:dist_vectors}
\end{figure}

In this context, focal regions can be defined as clusters of objects which are obtained with a Gaussian mixture model that takes the location information of ground-truth (GT) boxes as input. The location information consists of a vector of bounding box distances to grid of evenly sampled points in the image as seen in Figure~\ref{fig:dist_vectors}. This method yields better results compared to directly using coordinates of boxes.

Number of focal regions is selected depending on the number of GT boxes for the respective image. Number of focal regions ($N_f$) can be written as:

\begin{equation}\centering \label{nfocal1}
\centering
N_f = log_2(N_{gt})+2
\end{equation}
where $N_{gt}$ is the number of GT boxes. Let $\va{x}$ be a $1\times M$ sized distance vector of the $i^{th}$ GT box in an image, and $\boldsymbol{X}$ be the  $N_f  \times M $ sized array of feature vectors. Gaussian Mixture Model can be defined as:
\begin{equation}\centering \label{nfocal2}
\centering
\begin{split}
&p_{\va{x}} = \sum_{j=1}^{N_f}\phi_j \mathcal{N}(\va{x}|\va{\mu_j},\Sigma_j),\\
&\sum_{j=1}^{N_f}\phi_j = 1
\end{split}
\end{equation}
where $\mu_j$ and $\sigma_j$ are the mean and variance of the $j^{th}$ cluster.

Expectation maximization algorithm is used to fit the model. Once the EM algorithm has run to completion, the fitted model can be used to perform clustering on GT bounding boxes. Given the model's parameters, probability that a GT bounding box belongs to a cluster is calculated as:

\begin{equation}\centering \label{gmmpred}
\centering
p_{C_i|\va{x}} = \frac{\phi_i \mathcal{N}(\va{x}|\va{\mu_i},\Sigma_i)}{\sum_{j=1}^{N_f}\phi_j \mathcal{N}(\va{x}|\va{\mu_j},\Sigma_j)}
\end{equation}

\begin{figure*}[t!]
\begin{center}
   \includegraphics[width=1\linewidth]{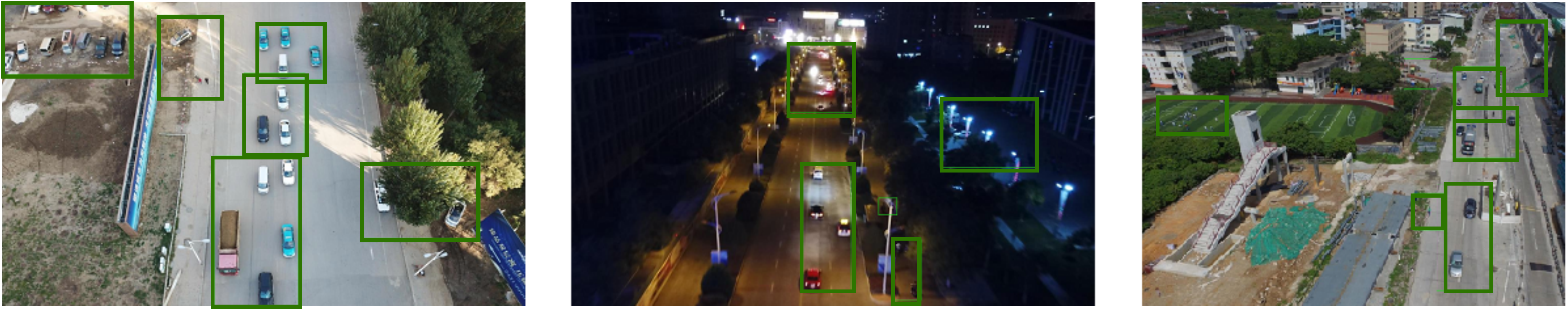}
\end{center}
   \caption{ Focal region ground truth examples. For each image, we fit a Gaussian Mixture Model. From generated clusters, target bounding boxes of focal regions are obtained.
    }
\label{fig:examples}
\end{figure*}

After calculation of clusters, focal regions are selected as the minimum sized box that includes all bounding boxes with a $20$ pixel gap on each side in respective cluster. Because of the gap, there might be truncated objects in focal regions. Generated focal regions are used as ground truth bounding boxes for focus stage as seen in Figure~\ref{fig:examples}.

\subsection{Detection Stage}
After obtaining the focal regions, a dedicated detector is utilized to perform object detection on these regions. Obtained regions are resized to a higher resolution. This approach improves performance of small object detection. 

In this stage, Generalized Focal Loss (GFL) is adopted as the base detector. Backbone of the model is selected as ResNeXt-101 network with deformable convolutional layers. On the neck,  Feature pyramid network (FPN) is used to improve detection performance by using features from different stages, and the last part is the detection head of the model which predicts bounding boxes of objects. The deformable convolution layer is used in the last three stages. The deformable convolution yields better results than traditional convolutional layers at detecting small objects, as it is able to dynamically change its receptive field and improves the detection performance which is hindered by geometric transformations.

On detection stage, focal regions that are obtained with GMM are cropped and resized to gather a new set of data. GT bounding boxes are obtained and refined to focal region crops. Truncated GT boxes are included if at least $30 \%$ of the box resides within the cropped region.


\subsection{Post Processing}
To obtain final predictions of object bounding boxes, predictions from detection stage must be merged as model outputs predictions of focal regions. The post processing steps that are applied to improve performance consist of Incomplete Box Suppression (IBS) and Non-Max Suppression (NMS).

\subsubsection{Incomplete Box Suppression}
 Models that are utilizing region search have certain problems. For instance, merging detections of target regions might be difficult, as there might be overlapped regions and truncated objects. This problem yields multiple bounding box predictions on same object. Because of the truncation, predicted bounding boxes are not fully overlapped. Thus, non-max suppression is not able to suppress these kind of false predictions. However, these predictions decrease the AP score.

\begin{figure}[h!]
\begin{center}
   \includegraphics[width=0.85\linewidth]{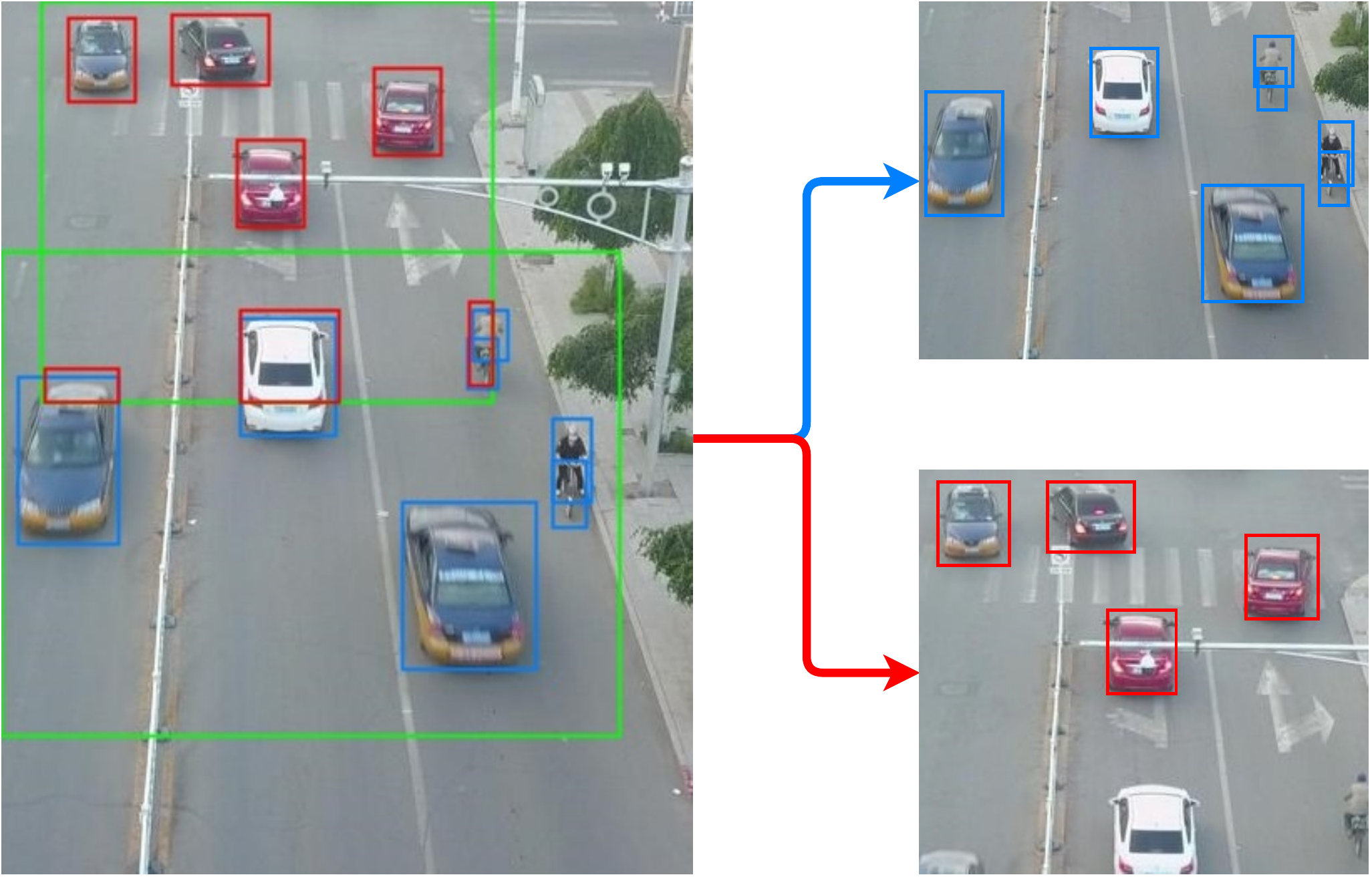}
\end{center}
   \caption{ Incomplete Box Suppression (IBS). Overlapping blue bounding boxes suppress the incomplete red ones and decreases the false positives. 
    }
\label{fig:ibs}
\end{figure}
In general, non-max suppression is used to eliminate highly overlapping boxes. It works well enough for traditional object detection problem. However, in most of the region search approaches, there is a final step which is merging the predictions of target regions. This creates a new problem. Overlapped regions and truncated objects in these regions which lowers the overall performance as detector might predict a full version of bounding box, and a truncated version of the bounding box for the same object as shown in Figure~\ref{fig:ibs}. Generally, intersection over union of these bounding boxes are small. Thus, they are able to escape from NMS. Truncated objects are also a problem on their own. False class predictions are common for truncated objects. As a result, false positives increase and AP score decreases. Incomplete Box Suppression (IBS) is proposed to reduce these kind of problems. 

Essentially, IBS has the same principle with the NMS algorithm: finding the overlapped bounding boxes, selecting the box with the highest confidence value, and suppressing the others. While NMS uses a simple Intersection over Union(IoU) threshold  to find overlaps, in the IBS, overlapping focal regions and object bounding boxes are both utilized to decide which box to suppress.

Let $C_i$ and $B_{ij}$ be the $i^{th}$ focal region coordinates and the $j^{th}$ box coordinates in that region.
\begin{itemize}
  \item First step is to calculating IoU between focal region $C_i$ and other focal regions to find $C_i$'s overlaps. Overlapping focal regions are obtained after applying a threshold to calculated IoUs.
  \item Second step is to clip objects box coordinates in overlapping focal regions to the $i^{th}$ focal region's coordinates and gather the boxes whose areas are greater than zero.
  \item Final step is to calculate IoU between clipped boxes and  the $B_{ij}$. If any of the IoU scores are greater than the selected threshold, the $B_{ij}$ is suppressed.
\end{itemize}

The IoU threshold for focal regions is experimentally selected as $0.05$, and IoU threshold for bounding boxes is, again, experimentally selected as $0.5$.

\subsubsection{Non-max Suppression}
Non-max suppression is applied to suppress overlapped detections after the merging of focal regions. Some of the overlapped focal regions contain same objects which causes duplicate box predictions. To mitigate this behavior, boxes with the highest confidences are selected and other boxes are suppressed. Intersection over union threshold for NMS is selected as $0.5$.

\section{Experimental Results}
\subsection{Implementation Details}
We implement Focus\&Detect based on the publicly available MMDetection \cite{mmdetection} and Pytorch. Generalized Focal Loss with Feature Pyramid Network is selected as the base detectors for both focus and detection stages. ResNet-50 and ResNeXt-101 are utilized as feature extraction networks in Focus stage and Detection stage, respectively. Focal detections are merged using NMS and IBS to obtain final predictions.
\paragraph{Training phase} The input size of Focus stage is randomly sampled from $400\times 1400$ to  $1200\times 1400$ in each step and samples are uniformly distributed for VisDrone dataset \cite{zhu2018vision}. \textcolor{black}{For UAVDT dataset \cite{du2018unmanned}, the input sizes of Focus stage and Detect stage are randomly sampled from $400\times 1000$ to  $800\times 1000$ and $400\times 800$ to  $800\times 800$, respectively.}  Flip augmentation is used with probability of $0.5$. For both Focus and Detection models, gradient descent with momentum, weight decay and learning rate scheduling is used. Both models are trained for 24 epochs. We set the initial learning rate to 0.01, at the 16th and 22th epoch, learning rate decreased to $0.001$ and $0.0001$. Beta parameter of momentum is selected as 0.9 for both models. Weight decay is applied with $0.0001$ ratio. Both model leverages Synchronized Batch Normalization and Group Normalization on the backbone and FPN, respectively. 
\paragraph{Testing phase} In the experiments with VisDrone dataset, the input sizes of Focusing and Detection models are selected as $1200\times 1400$ and $600\times 1000$. \textcolor{black}{On the other hand, in the experiments with UAVDT dataset the input sizes of Focusing and Detection models are selected as $600\times 1000$ and $600\times 800$.} While merging detections of focal regions, NMS and IBS are applied. IoU threshold for NMS is $0.5$. After NMS, IBS is applied to reduce false positives caused by truncated objects in focal regions. IoU thresholds for IBS are selected as $0.05$ and $0.5$, where first threshold is for overlapping focal regions and second threshold is for overlapping truncated objects in focal regions after clipping is applied.

\subsection{Dataset and Evaluation Metric}

In this work, we utilize VisDrone2021 Detection dataset \cite{zhu2018vision} which consists of 6,471 images for training, 548 images for validation and 3,190 images for testing. It is an aerial image dataset which is obtained by drone-mounted cameras taken from 14 cities in China. Moreover, this dataset has 10 classes of objects which are non-uniformly distributed. 

In order to assess the performance of our method, we use the evaluation protocol in MS COCO \cite{lin2014microsoft}. To be precise, we report AP, $AP_{50}$ and $AP_{75}$ scores, where AP shows the average precision for ten IoU thresholds whose range is from 0.5 to 0.95 and $AP_{50}$ is the average precision with the IoU threshold of 0.5. Similarly, $AP_{75}$ is the average precision with the IoU threshold of 0.75. In addition to these, we present $AP_{S}$, $AP_{M}$ and $AP_{L}$  scores, where $AP_{S}$, $AP_{M}$ and $AP_{L}$ represent the AP for objects with the area less than $32 \times 32$, less than $96 \times 96$ and larger than $96 \times 96$ pixels, respectively.

\textcolor{black}{We also make experiments on UAVDT dataset \cite{du2018unmanned}, which is an aerial image dataset consists of around 41k frames with 840k bounding boxes. It has car, truck, and bus categories. The class distribution of the UAVDT dataset is extremely imbalanced where the truck and bus classes cover less than 5\% of bounding boxes. Following the dataset authors' convention, we combine the three classes into one vehicle class and report PASCAL VOC AP score with an IoU threshold of 0.7 based on \cite{du2018unmanned}.}

\subsection{Results}

 We report results on VisDrone test set in Table~\ref{tab:testdev}, comparing our model to the baseline model. Focus\&Detect significantly improves performance  on  pedestrian, person, bicycle classes which are mostly consist of small objects. On the other hand, baseline model yields a better performance in mAP@50 score on tricycle and awning-tricycle classes.
 
To have a fair comparison with the reported performance results of other state-of-the-art techniques, we also present our results on VisDrone validation set in Table~\ref{tab:comparison}. We compare our method with the other state-of-the-art region search based methods for object detection, in terms of  number of images which are forwarded to detectors, evaluation metrics for precision and inference time per image.

The stated inference time results of our method and CRENet~\cite{wang2020object} are obtained on RTX 2080 Ti, while others are obtained on GTX 1080 Ti GPUs. 
We report the average inference time per image, since number of focal regions per image differs, similar to SAIC-FPN \cite{zhou2019scale}.

\begin{table*}[h!]
    \begin{center}
    \caption{Comparison between base model GFL and Focus\&Detect, in terms of class-wise AP@50[\%] scores on VisDrone \textbf{test-dev} dataset. \textcolor{black}{Results indicate that the proposed Focus\&Detect method significantly improves AP scores of the classes with the largest number of small objects, namely, "Pedestrian", "Person", and "Bicycle". In the experiments, GFL is tested in $2160 \times 3840 $ resolution with flip augmentation, Focus\&Detect is tested in $600\times1000$ resolution without any test-time augmentation.} The highest AP scores for each column are printed in boldface.}
    \label{tab:testdev}
    \resizebox{\textwidth}{!}{\begin{tabular}{|l|c|c|c|c|c|c|c|c|c|c|c|}
    \hline
         \multirow{2}{*}{Method}
          & All  & \multirow{2}{*}{Pedestrian} & \multirow{2}{*}{Person} & \multirow{2}{*}{Bicycle} &\multirow{2}{*}{Car} & \multirow{2}{*}{Van} & \multirow{2}{*}{Truck} & \multirow{2}{*}{Tricycle} & Awning- & \multirow{2}{*}{Bus} & \multirow{2}{*}{Motor}  \\ 
           &classes&&&&&&&& tricycle && \\
         \hline \hline

         GFL & 52.0 & 52.8 & 34.1 & 30.3 & \textbf{86.4} & 56.1 &60.2 &\textbf{39.1}&\textbf{34.9}&69.6& \textbf{56.3} \\
         Focus\&Detect & \textbf{52.6} &\textbf{56.5} & \textbf{38.6} & \textbf{32.1} & 86.1 & \textbf{56.3} & \textbf{60.8} & 36.8 & 32.6 & \textbf{70.2}& 56.1 \\
         \hline
    \end{tabular}}
    \end{center}
\end{table*}

\begin{table*}[t!]

\begin{center}
\caption{Results on VisDrone validation set. Our method is compared with the other region search based object detection methods.
"\#img" is the number of images that the detectors are fed by, whereas the last column shows the inference time per image in seconds. 
We report the average inference time per image for our method, since number of focal regions per image differs.
"SS" shows single-scale inference.
* represents the result of EIP with the detector of Faster R-CNN and FPN, obtained by \cite{yang2019clustered} which means partition image equally into six non-overlapping pieces to obtain regions to be focused. Bold values represent the column-wise best scores.
}
\label{tab:comparison}
\resizebox{\textwidth}{!}{\begin{tabular}{|l|c|c|c|c|c|c|c|c|c|c|}
\hline

\multirow{2}{*}{
Method} & \multirow{2}{*}{Backbone} & Image & \multirow{2}{*}{\#Img} & \multirow{2}{*}{$AP [\%]$} & \multirow{2}{*}{$AP_{50} [\%]$ }& \multirow{2}{*}{$AP_{75} [\%]$} & \multirow{2}{*}{$AP_{S} [\%]$} & \multirow{2}{*}{$AP_{M} [\%]$} & \multirow{2}{*}{$AP_{L} [\%]$} & \multirow{2}{*}{s / img}  \\
&&Resolution&&&&&&&&\\
\hline\hline

EIP* & {ResNeXt-101} & {$600 \times 1000$} & {3288} & {24.4} & {47.8} & {21.8} & {17.8} & {34.8} & {34.3} & {0.936} \\
ClusDet (SS) \cite{yang2019clustered} & ResNeXt-101 & $600 \times 1000$ & 2716 & 28.4 & 53.2 & 26.4 & 19.1 & 40.8 & 54.4 & 0.773 \\ 
DMNet \cite{li2020density} & ResNeXt-101 & $600 \times 1000$ & 2736 & 29.4 & 49.3 & 30.6 & 21.6 & 41.0 & 56.9 & - \\
CRENet \cite{wang2020object} & Hourglass-104 & $1024 \times 1024$ & 2337 & 33.7 & 54.3 & 33.5 & 25.6 & 45.3 & \textbf{58.7} & 0.901 \\
SAIC-FPN \cite{zhou2019scale} & ResNeXt-101 & - & - & 35.7 & 63.0 & 35.1 & -& - & - & 0.252 $\sim$ 2.568 \\
AdaZoom \cite{xu2021adazoom} & ResNeXt-101 & - & - & 40.3 & \textbf{66.9} & 41.8 & - & - & - & -\\

\textbf{Focus\&Detect} &ResNeXt-101&$600 \times 1000$ & 9004 & \textbf{42.0} & 66.1 &\textbf{44.6}&\textbf{32.0}&\textbf{47.9} & 54.5 & 1.362\\ 

\hline
\end{tabular}}
\end{center}
\end{table*}

\begin{table*}[t!]
\begin{center}
\caption{Comparison of our method with other state-of-the-art methods for object detection on VisDrone validation set, where we report the results on original papers for other methods. Bold values represent the column-wise best scores. 
}
\label{tab:SOTA}
\resizebox{\textwidth}{!}{\begin{tabular}{|l|c|c|c|c|}
\hline
  Method         & Backbone & $AP [\%]$ & $AP_{50} [\%]$  & $AP_{75} [\%]$    \\
\hline\hline
  RRNet~\cite{chen2019rrnet} &Hourglass & 32.92 & -  & 31.33 \\
  CRENet~\cite{wang2020object}  & Hourglass-104 &33.70  & 54.30  & 33.50  \\
  DMNet~\cite{li2020density} & ResNet-50   & 28.20  &  47.60 & 28.90  \\
  CascadeNet~\cite{zhang2019dense} &ResNet-50  &30.12  &58.02 & 27.53 \\
  GLSAN~\cite{deng2020global}     &ResNet-50   &30.70  &55.40  & 30.00   \\
  SAMFR~\cite{wang2019spatial} &ResNet-50  &33.72  &58.62 & 33.88 \\
  AdaZoom (w/Faster R-CNN)\cite{xu2021adazoom}  & ResNet-50    & 36.19  & 63.50  & 36.11 \\
  MPFPN~\cite{liu2020small2}  &ResNet-101 & 29.05 & 54.38 & 26.99  \\
  GLSAN~\cite{deng2020global} & ResNet-101 &30.70  & 55.60  & 29.90 \\
  ClusDet~\cite{yang2019clustered} & ResNeXt-101 &32.40  &56.20  &31.60 \\
  QueryDet~\cite{yang2021querydet} & ResNeXt-101 & 33.91 & 56.12 & 34.85 \\
  SAIC-FPN~\cite{zhou2019scale} &ResNeXt-101 & 35.69 &62.97 &35.08 \\
  AdaZoom (w/Faster R-CNN) \cite{xu2021adazoom} & ResNeXt-101 & 37.58 & 66.25 & 37.34 \\
  AdaZoom (w/Cascade R-CNN)  \cite{xu2021adazoom} & ResNeXt-101 & {40.33}  & \textbf{{66.94}}  & {41.77}  \\
  DREN~\cite{zhang2019fully} & ResNeXt-152 &30.30 &-&- \\
  \textbf{Focus\&Detect} & ResNeXt-101 & \textbf{42.06} & 66.12 & \textbf{44.64} \\

\hline
\end{tabular}}
\end{center}
\end{table*}

Results show that evenly image partition (EIP) method obtains the worst results \cite{yang2019clustered}, despite the high number of regions generated where it can be defined as the most straightforward approach among the region search based methods.
Moreover, our method obtains 42.06 AP score on VisDrone validation set which outperforms other region search based object detection methods, reported in the literature.
Although we set lower image resolution values as compared with CRENet, our method clearly surpasses CRENet in terms of $AP_S$ which happens to be the reported best result for $AP_S$.

Furthermore, we compare our method with the state-of-the-art methods based on various techniques, on VisDrone validation set as presented in Table~\ref{tab:SOTA}. Results indicate that our method achieves 42.06 $AP$ and 44.64 $AP_{75}$ scores which are the best scores among state-of-the-art methods, reported in the literature.

\textcolor{black}{We report results on UAVDT dataset in Table~\ref{tab:UAVDT}, comparing our model to the state-of-the-art reported methods on UAVDT dataset \cite{du2018unmanned}. Compared to other methods Focus\&Detect significantly improves performance. }
\begin{table*}[t!]
\begin{center}
\caption{Comparison of our method with other state-of-the-art methods for object detection on UAVDT test set, where we report the results on original papers for other methods. 
}
\label{tab:UAVDT}
\resizebox{4.5cm}{!}{\begin{tabular}{|l|c|c|c|c|}
\hline
  Method         &  $AP [\%]$   \\
\hline\hline
  RetinaNet~\cite{lin2017focal}  & 33.95 \\
  LRF-Net~\cite{wang2019learning}  & 37.81 \\
  FPN~\cite{lin2017feature}  & 49.05 \\
  NDFT~\cite{wu2019delving}  & 52.03 \\
  Focus\&Detect  & \textbf{54.16} \\

\hline
\end{tabular}}
\end{center}
\end{table*}
\textcolor{black}{Results show that our framework outperforms FPN~\cite{lin2017feature} and RetinaNet~\cite{lin2017focal}. UAVDT dataset \cite{du2018unmanned} does not solely focus on small objects. Contrary to VisDrone dataset, there are no annotation for very small objects and they are ignored in test time. Nevertheless, Focus\&Detect surpasses the methods that are reported in the literature. }
\subsection{Ablation Study}
\begin{figure*}
\begin{subfigure}{.99\textwidth}
  \centering
  \includegraphics[width=0.8\linewidth]{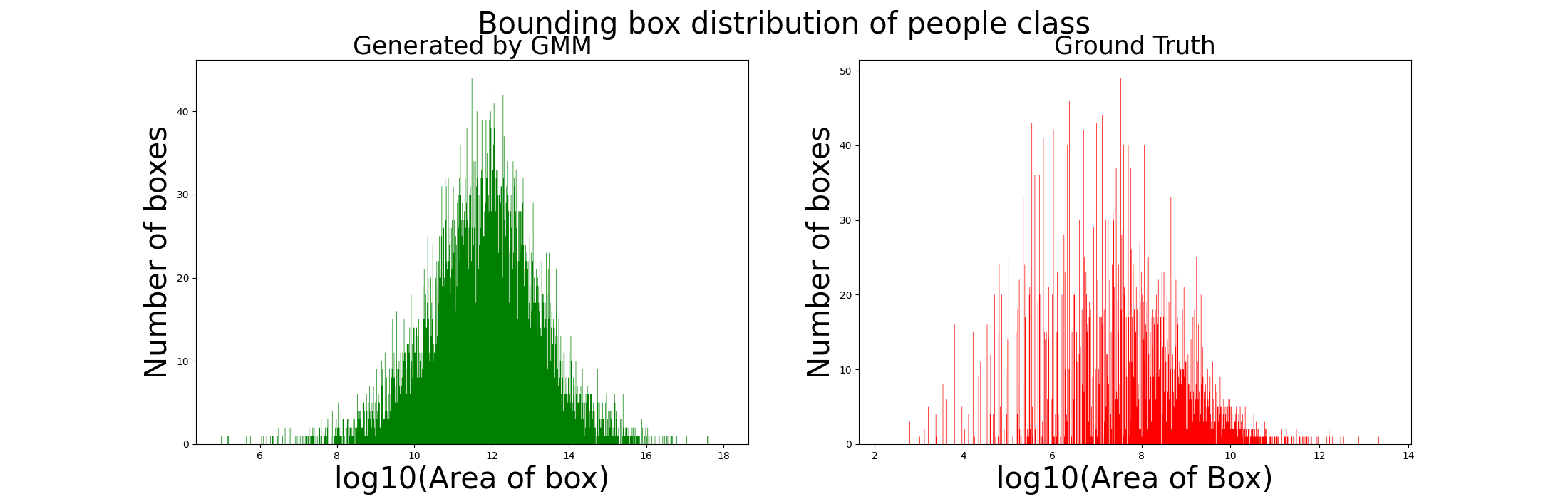}  
  \caption{}
  \label{fig:sub-first}
\end{subfigure}
\newline
\begin{subfigure}{.99\textwidth}
  \centering
  \includegraphics[width=0.8\linewidth]{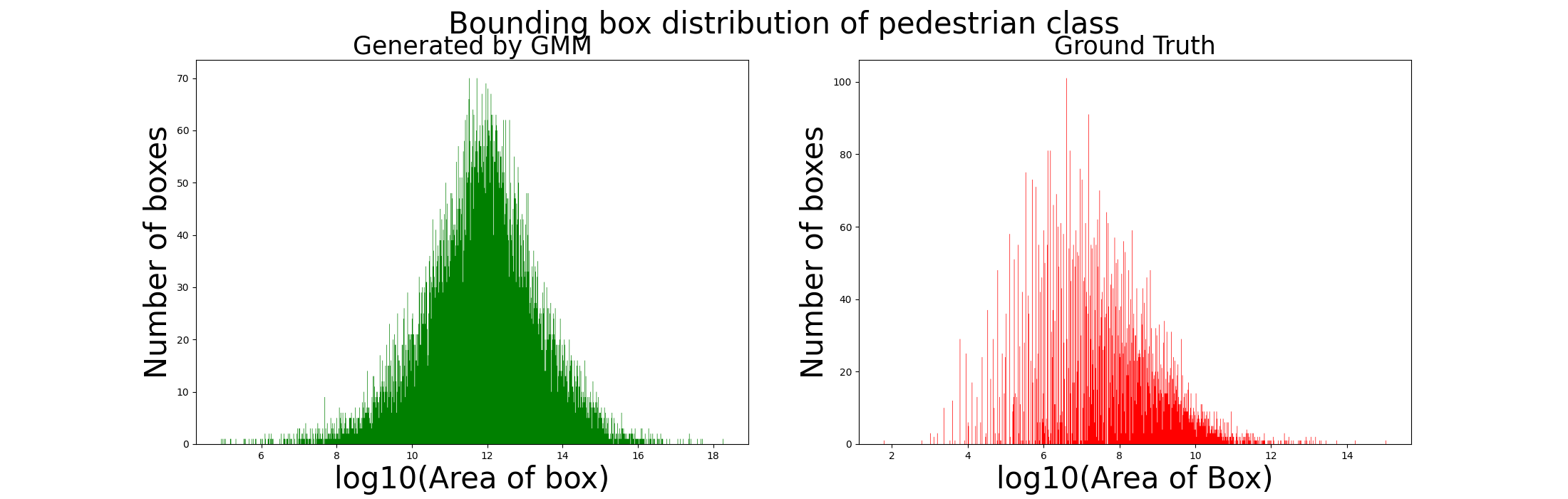}  
  \caption{}
  \label{fig:sub-second}
\end{subfigure}


\begin{subfigure}{.99\textwidth}
  \centering
  \includegraphics[width=0.8\linewidth]{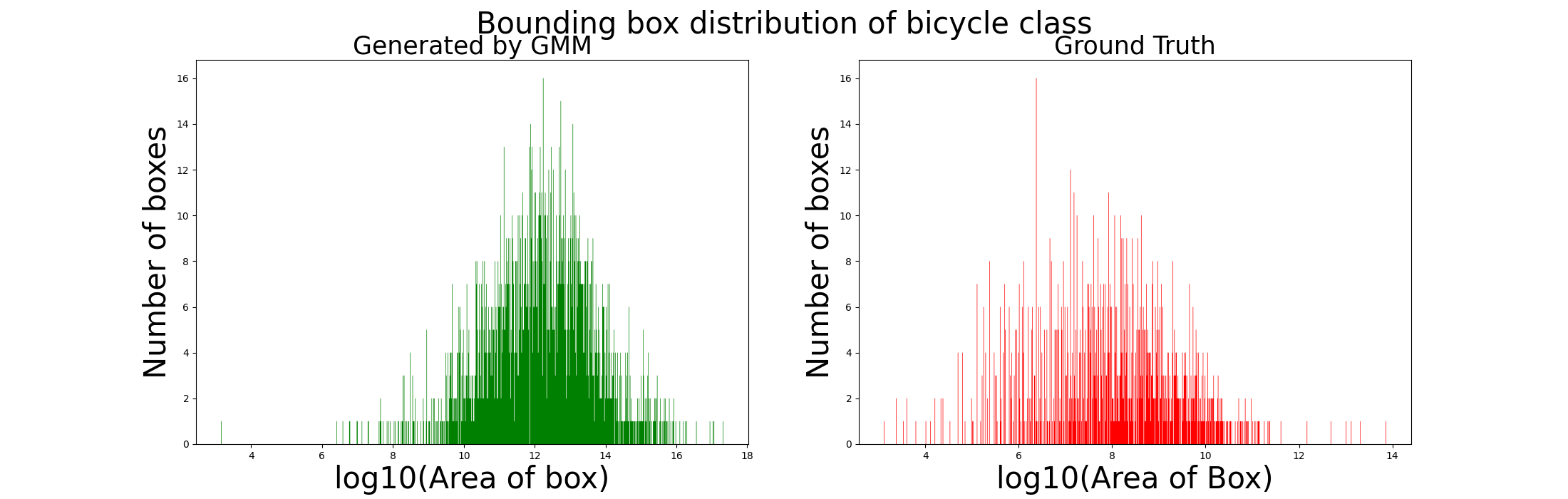} 
  \caption{}
  \label{fig:sub-third}
\end{subfigure}
\newline
\begin{subfigure}{.99\textwidth}
  \centering
  \includegraphics[width=0.8\linewidth]{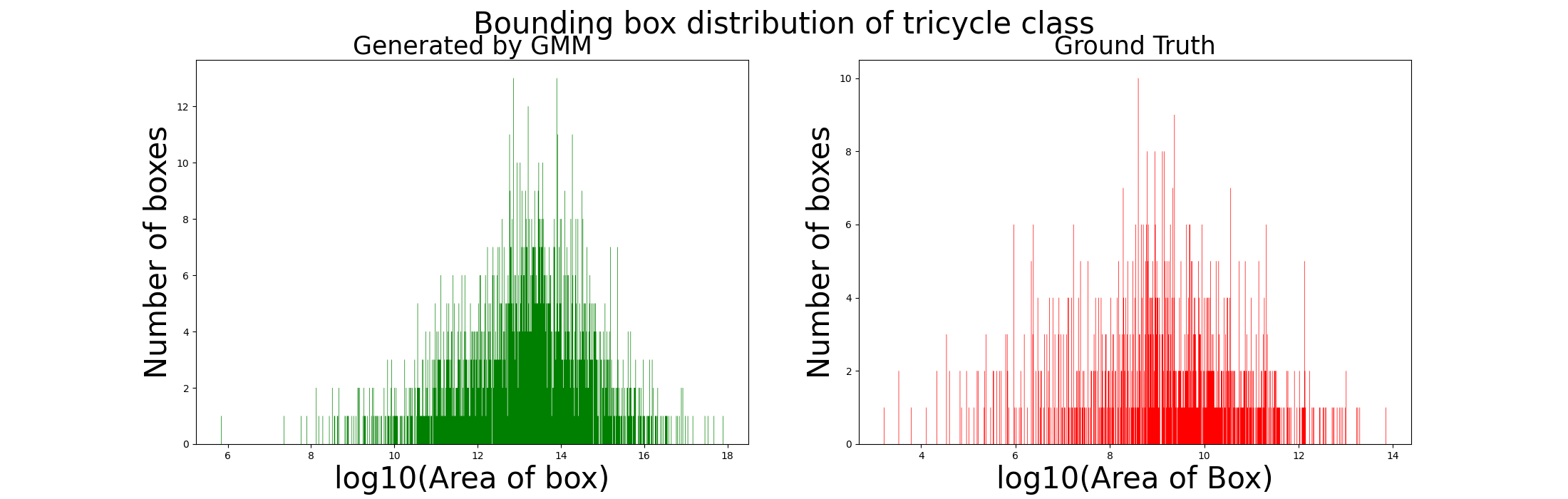}  
  \caption{}
  \label{fig:sub}
\end{subfigure}
\caption{Effect of GMM clustering: We compare the areas of ground truth boxes and ground truth box areas of clusters obtained from GMM. We can clearly see that mean area is increased for each class. Moreover, distribution of boxes became normally distributed which clearly improve detection performance of models. (a,b,c,d) Distribution of bounding boxes for each class leverages from GMM clustering. Despite sizes of object clusters  differ, bounding box areas are normalized. }
\label{fig:gmmclusteringeffect}
\end{figure*}
\begin{table*}[h!]
    \begin{center}
    \caption{Comparison between base model GFL, Focus\&Detect with IBS and Focus\&Detect without IBS, in terms of class-wise AP@50[\%] scores. GFL is tested in $2160 \times 3840 $ resolution, Focus\&Detect is tested in $600\times1000$ resolution. }
    \label{tab:ablation}
    \resizebox{\textwidth}{!}{\begin{tabular}{|l|c|c|c|c|c|c|c|c|c|c|c|}
    \hline
         \multirow{2}{*}{Method}
          & All  & \multirow{2}{*}{Pedestrian} & \multirow{2}{*}{Person} & \multirow{2}{*}{Bicycle} &\multirow{2}{*}{Car} & \multirow{2}{*}{Van} & \multirow{2}{*}{Truck} & \multirow{2}{*}{Tricycle} & Awning- & \multirow{2}{*}{Bus} & \multirow{2}{*}{Motor}  \\ 
           &classes&&&&&&&& tricycle && \\
         \hline \hline

         GFL & 62.8 & 74.4 & 61.4 & 48.5 & 90.6 & 64.1 &57.8 &52.7&29.9&77.6 & 71.4 \\
         F\&D w/o IBS & 63.8 & 75.5 & 61.7 & 49.7 & 90.4 & 65.2 & 58.9 & 55.5 & 30.3 & 80.5 &69.9 \\
         F\&D w/ IBS & 66.1 & 78.6 & 67.6 & 53.1 & 91.7 & 67.2 & 60.7 & 57.5 & 32.0 & 81.8 & 73.2 \\
         \hline
    \end{tabular}}
    \end{center}
\end{table*}

We conduct ablation experiments in order to validate the contributions of GMM and IBS to the overall performance of object detection. To clarify the contributions, we report the class-wise results on VisDrone validation set in Table~\ref{tab:ablation} and show the effect of GMM clustering in Figure \ref{fig:gmmclusteringeffect}. 

\paragraph{Effect of GMM Clustering}
Aerial images contain different sized objects depending on angle and altitude of the drone. Training an object detection model for aerial images is challenging as the data distribution does not contain all 
the different scales of objects. If ground truth bounding box areas are well distributed over real data distribution, object detector yields higher detection performance. 

As seen in Figure \ref{fig:gmmclusteringeffect},  clustering objects has a normalizing effect on areas of bounding boxes. GMM clustering increases the mean box area which helps to improve performance on small objects.

We can compare effect of GMM clustering with resizing image to high resolution and multi-scale training. Increasing resolution shifts the mean of objects size. However, it does not normalize the bounding box areas. On the other hand Multi-scale training has similar effects, as it shifts means of bounding box areas and normalizes the box areas.

\paragraph{Effect of IBS}

Results point out that F\&D without IBS method improves particularly detection scores of `bus' and `tricycle' classes compared to GFL method. 

In addition to this, proposed IBS method advances the detection performance of all classes by 2.3\%.
Moreover, IBS provides a performance boost on all classes, especially on classes of small objects such as `person', `bicycle', `motor' and `pedestrian'.

\section{Conclusion}
A two stage framework is proposed to solve small object detection problem in aerial images. 
The proposed method is region search based where we utilize a Gaussian Mixture Model to generate focal regions for object detection. GMM method has a normalization effect on GT box sizes as cropping and resizing the image to a fixed resolution relatively forces objects to an average size for each class. We also propose the Incomplete Box Suppression (IBS) method to mitigate the truncated box problem that arise while merging the target regions.

Results show that the proposed IBS method improves the detection performances of all classes, especially of small object classes. \textcolor{black}{GMM clustering normalizes the object scales across regions and increases overall performance.} \textcolor{black}{Furthermore, our method achieves the state-of-the-art performance on VisDrone validation set and UAVDT test set comparing to other small object detection methods, reported in the literature.} Moreover, our method obtains the best $AP_S$ score among all other methods, which indicates the positive impact of the proposed framework on small object detection.

\section{Funding}
This work was supported in part by ASELSAN Inc. with grant number 65834 STB and in part by the Scientific and Technical Research Council of Turkey, TÜBİTAK, with grant number 121E378.
\section*{CRediT authorship contribution statement}
\textbf{Onur Can Koyun:} Conceptualization, Methodology, Software, Writing - original draft. \textbf{Reyhan Kevser Keser:} Conceptualization, Methodology, Analysis and interpretation of data, Writing - original draft. \textbf{İbrahim Batuhan Akkaya:} Supervision, Writing - review \& editing. \textbf{Behçet Uğur Töreyin:} Supervision, Writing - review \& editing

\bibliographystyle{unsrt}  
\bibliography{references.bib}

\end{document}